\documentclass[aps,pre,reprint,amsmath,amssymb]{revtex4-2}
\usepackage{txfonts}
\usepackage{algorithmic}
\usepackage{algorithm}
\usepackage{graphicx}
\usepackage{bm}
\usepackage{url}

\newcommand{\argmin}{\mathop{\rm arg~min}\limits}

\newcommand{\1}{\mbox{1}\hspace{-0.25em}\mbox{l}}

\begin{document}

\title{Compression of the Koopman matrix for nonlinear physical models via hierarchical clustering}

\author{Tomoya Nishikata and Jun Ohkubo}

\affiliation{Graduate School of Science and Engineering, Saitama University, Sakura, Saitama, 338--8570 Japan}

\begin{abstract}
Machine learning methods allow the prediction of nonlinear dynamical systems from data alone. The Koopman operator is one of them, which enables us to employ linear analysis for nonlinear dynamical systems. The linear characteristics of the Koopman operator are hopeful to understand the nonlinear dynamics and perform rapid predictions. The extended dynamic mode decomposition (EDMD) is one of the methods to approximate the Koopman operator as a finite-dimensional matrix. In this work, we propose a method to compress the Koopman matrix using hierarchical clustering. Numerical demonstrations for the cart-pole model and comparisons with the conventional singular value decomposition (SVD) are shown; the results indicate that the hierarchical clustering performs better than the naive SVD compressions.
\end{abstract}

\maketitle

\section{Introduction}

Physical simulation is used in many research fields, for example, to predict natural phenomena, design industrial robots, and simulate the behavior of objects in games. While we perform numerical simulations based on system equations, machine learning methods are also available for the physical simulations. The simulations based on machine learning do not require the system equations; only a training data set is enough to predict the system behavior. Many systems have nonlinearity, and methods based on neural networks are hopeful candidates for practical cases. Of course, the neural networks are also nonlinear, and the computational cost is high. 

Recently, several works focused on the analysis based on the Koopman operator theory \cite{Koopman1931}, in which we employ linear analysis even if the underlying system is nonlinear. Instead of the nonlinear computation for the original system equations for the state, we consider the observable space in the Koopman operator theory. The Koopman operator acts on the observable function linearly, and we can employ various methods in linear algebra. However, the observable space is a function space and infinite-dimensional. Hence, we must approximate the infinite-dimensional Koopman operator as a finite-dimensional Koopman matrix in practical computation. One of the methods to construct the Koopman matrix from data is the dynamic mode decomposition (DMD) \cite{Schmid2010}. An extension of the DMD is the extended dynamic mode decomposition (EDMD) \cite{Williams2015}, in which we introduce a dictionary and achieve more accurate predictions. There are some extensions of DMD and EDMD; DMDc (DMD with control) and EDMDc (EDMD with control) \cite{Proctor2016} can deal with cases with external control inputs. Recently, the Koopman operators using the EDMD have been extensively studied for the prediction and control of physical models \cite{Bruder2021,Kaiser2021}. Furthermore, various applications and studies of Koopman matrices have been discussed, for example, to analyze power systems \cite{Susuki2016,Takamichi2022}, reduce computational time in deriving Koopman matrices \cite{Li2022}, add robustness against noise \cite{Dicle2016}, and construct Koopman matrices using prior knowledge of the underlying model \cite{Folkestad2020,Li2019}. For the Koopman operator theory, see the recent review \cite{Brunton2022}.

While the Koopman operators are available for various purposes, there is a problem with practical usage; the size of the dictionary becomes large for higher-dimensional systems. Hence, the computational complexity is enormous. Of course, one could construct the dictionary with the aid of neural networks; in \cite{Li2019}, the authors employed not only the conventional basis functions but also the dictionary based on the neural networks. In \cite{Terao2021}, the dictionary learning with neural ordinary differential equations was discussed. However, it will be beneficial to avoid the usage of neural networks because of their high computational costs. In addition, robotics applications would avoid the black-box natures of neural networks. Hence, it is preferable to employ simple dictionary functions if possible.

Once one fixes the dictionary, we obtain the Koopman matrix explicitly. Then, it is also crucial to reduce the computational costs related to the Koopman matrix. In \cite{Li2019}, the authors discussed the reduction of computational cost in the construction stage of the Koopman matrix. One can combine the prior knowledge of the underlying model with the Koopman matrix; in \cite{Li2019}, it is possible to reduce the learning cost by making some elements of the Koopman matrix common. Although this approach reduces the computational costs for the learning process, the use of the Koopman matrix in the prediction stage was the same as the conventional ones in \cite{Li2019}. Then, how about the methods to reduce computational cost in the prediction steps?

In the present paper, we propose a method to compress the constructed Koopman matrix. The proposed method employs hierarchical clustering to extract similar rows and columns. Based on the clustering results, we perform the grouping of similar rows and columns, which compresses the Koopman matrix and the dictionary. Since the compressed Koopman matrix is not square, it is unavailable for calculating the time evolution repeatedly. Then, we introduce an additional matrix to recover the size. As a demonstration, we perform numerical experiments on a data set generated from a cart-pole model; we also compare the numerical results with a conventional matrix compression with the singular value decomposition (SVD).

The outline of the present paper is as follows. In Sec.~2, we briefly review the Koopman operator theory and the EDMD algorithm. The main proposal is yielded in Sec.~3; the compression method based on hierarchical clustering is explained. In Sec.~4, we numerically demonstrate the proposed method. As an example, a data set generated from the cart-pole model is used. Section~5 gives some concluding remarks.

\section{Preliminaries}
\label{sec_preliminaries}

\subsection{A concrete example of physical model}

In this section, we briefly review the Koopman operator theory, the EDMD algorithm, and hierarchical clustering methods. It would be helpful to use a concrete example for readers' understanding, and then we sometimes use the cart-pole model as the concrete example. 

\begin{figure}
\begin{center}
    \includegraphics[width=3.5cm]{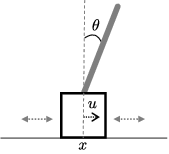}
    \caption{Cart-pole model. The external force $u$ only acts on the cart horizontally.}
    \label{fig:cart_pole}
\end{center}
\end{figure}

Figure~\ref{fig:cart_pole} shows the cart-pole model. The horizontal coordinate of the cart is $x$, and the cart can move only horizontally. The angle between the pole and the cart is $\theta$. The corresponding velocities are denoted as $(\dot{x},\dot{\theta})$. The state equation for the cart-pole model is nonlinear \cite{Mills2009}; since only a data set is enough to apply the Koopman operator theory, we do not need to know the state equation here. There is an external force $u$ that acts on the cart horizontally to control the pole to stand. However, the force does not appear in the following explanation because only a data set for the system states under proper control situations is employed. As explained in Sec.~4, we use the time-series data for the cart-pole model controlled by a reinforcement learning model.

\subsection{Koopman operator and the EDMD algorithm}

For the details of the EDMD algorithm, see \cite{Williams2015}. We here only review the necessary parts of the algorithm.

Consider a nonlinear time evolution function $F: \mathbb{R}^{D_{\mathrm{s}}} \to \mathbb{R}^{D_{\mathrm{s}}}$ in a $D_{\mathrm{s}}$-dimensional state space. The state at discrete time $t$ is denoted as $\bm{x}_t$, and then $F(\bm{x}_t)=\bm{x}_{t+1}$. In the cart-pole model, $D_{\mathrm{s}} = 4$ and the state is given as $\bm{x}_t = [x_t,\theta_t,\dot{x}_t,\dot{\theta}_t]$.

\begin{figure}[tb]
\begin{center}
    \includegraphics[width=8cm]{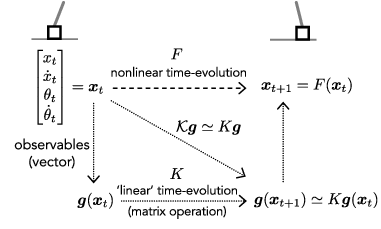}
    \caption{Linear computation via the Koopman operator. Instead of the original nonlinear dynamics, we employ the linear operator $\mathcal{K}$ in the observable space. In practice, we estimate a finite-dimensional matrix to approximate the linear operator.}
    \label{fig:Koopman}
\end{center}
\end{figure}

In the Koopman operator theory, we introduce an observable vector $\bm{g}(\bm{x})$ for the state $\bm{x}$. An observable, $g_i(\bm{x})$, is a function; for example, $g_i(\bm{x}) = \theta$ is a function to observe only the angle $\theta$. As depicted in Fig.~\ref{fig:Koopman}, a Koopman operator $\mathcal{K}$ is a linear operator that performs discrete-time evolution for the observable vector by linear computation. The discrete-time evolution equation using the Koopman operator is defined as follows:
\begin{align}
\mathcal{K}\bm{g}(\bm{x}_t) = \bm{g} \circ F(\bm{x}_t) = \bm{g}(\bm{x}_{t+1}),
\end{align}
where $\circ$ means the composition of functions. The Koopman operator allows linear computation for the nonlinear time evolution of the original system. Note that the observables are functions; hence, the Koopman operator acts on the corresponding function space. Since the function space is infinite-dimensional, we practically approximate the Koopman operator $\mathcal{K}$ as a finite-dimensional matrix, called a Koopman matrix $K$.

The EDMD algorithm is one of the algorithms to derive the finite-dimensional approximation of the Koopman operator from data \cite{Williams2015}. The EDMD algorithm requires snapshot pairs of time series data and a dictionary. The snapshot pairs are $s$ pairs of data before and after time evolution, i.e., $(\bm{X}_1=[\bm{x}_1,\bm{x}_2,\cdots,\bm{x}_s]$ and $\bm{X}_2=[\bm{x}_2,\bm{x}_3,\cdots,\bm{x}_{s+1}])$; we here simply make the snapshot pairs from a single time-series data. The dictionary is a vector of functions. Let $D$ be the size of the dictionary, i.e., the number of dictionary functions. Then, the dictionary $\bm{\psi}(\bm{x})$ is expressed as follows:
\begin{align}
    \label{Dictionary}
    \bm{\psi}(\bm{x}) = [\psi_1(\bm{x}),\psi_2(\bm{x}),\cdots,\psi_D(\bm{x})]^{\top}.
\end{align}
For example, monomial dictionary functions for the cart-pole model is written as
\begin{align*}
&\psi_1(\bm{x}) = 1,\,  \psi_2(\bm{x}) = x, \, \psi_3(\bm{x}) = \theta, \, \psi_4(\bm{x}) = \dot{x}, \, \psi_5(\bm{x}) = \dot{\theta}, \nonumber \\
& \psi_6(\bm{x}) = x^2, \, \psi_7(\bm{x}) = x \theta, \, \psi_8(\bm{x}) = x \dot{x}, \, \psi_9(\bm{x}) = x \dot{\theta}, \,\, \dots.
\end{align*}
It is easy to see that the dictionary size $D$ becomes large for high-dimensional systems. There are other dictionary functions, such as radial basis functions and neural network-based functions. However, the monomial dictionary functions are sometimes beneficial because of their simplicity and understandability. In the present paper, we employ the monomial dictionary functions. 

An observable function $g_i(\bm{x})$ is expressed in terms of the dictionary $\bm{\psi}(\bm{x})$:
\begin{align}
    \label{GandD}
    g_i(\bm{x}) = \sum^D_{d=1} a_{i,d} \psi_d(\bm{x}),
\end{align}
where $\{a_{i,d}\}$ are the expansion coefficients. Then, the action of the Koopman operator on the observable function yields
\begin{align}
    \label{KoopmanGandD}
    \mathcal{K} \circ g_i(\bm{x}) = \sum^D_{d=1} a_{i,d} (\mathcal{K} \circ \psi_d(x)).
\end{align}
The action of $\mathcal{K}$ on the observable vector $\bm{g}$ is defined as the element-by-element action. Then, Eq.~\eqref{KoopmanGandD} allows us to derive the time evolution of the observable vector $\bm{g}(\bm{x})$ by applying the Koopman operator to the dictionary $\bm{\psi}(\bm{x})$. Note that Eqs.~\eqref{GandD} and \eqref{KoopmanGandD} yield only approximations if the number of dictionary function is not enough.

The action of the Koopman operator $\mathcal{K}$ is approximated via the corresponding Koopman matrix $K$ as follows:
\begin{align}
    \mathcal{K} \circ \bm{\psi}(\bm{x}) \simeq K \bm{\psi}(\bm{x}).
\end{align}
To obtain the Koopman matrix, the conventional least-squares method is available. From the snapshot pairs ${\bm{X}_1,\bm{X}_2}$ and the dictionary $\bm{\psi}(\bm{x})$, we construct the data matrix $\Psi(\bm{X}_1) \in \mathbb{R}^{D\times s}$ and $\Psi(\bm{X}_2) \in \mathbb{R}^{D\times s}$. Then, the Koopman matrix is estimated via
\begin{align}
    K = \argmin_{\widetilde{K}} \| \Psi(\bm{X}_2) -  \widetilde{K} \Psi(\bm{X}_1)\|^2.
\end{align}
As derived in \cite{Williams2015}, the solution is given as
\begin{align}
    K = A G^{+},
\end{align}
where
\begin{align}
    A &= \frac{1}{s} \sum^s_{i=1} \bm{\psi}(\bm{x}_i) \bm{\psi}(\bm{x}_{i+1})^{\top}, \qquad
    G = \frac{1}{s} \sum^s_{i=1} \bm{\psi}(\bm{x}_i) \bm{\psi}(\bm{x}_i)^{\top}, 
\end{align}
and $G^{+}$ is the pseudo-inverse matrix of $G$.

\subsection{Hierarchical clustering}

\begin{figure}[t]
\begin{center}
    \includegraphics[width=5cm]{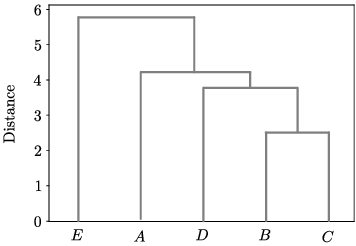}
    \caption{An example of a dendrogram constructed from hierarchical clustering. Here, the five elements are clustered hierarchically according to distance.}
    \label{fig:Dendrogram}
\end{center}
\end{figure}

Hierarchical clustering \cite{Murtagh2012,Murtagh2017} is one of the well-known clustering methods for unsupervised learning. To seek a hierarchy of clusters, we employ a bottom-up approach in this work. We merge the clusters with the smallest distance from each other and repeat this process. The stepwise classification yields a dichotomous tree called a dendrogram. An example of the dendrogram is shown in Fig.~\ref{fig:Dendrogram}; there are five elements (or clusters) $[A,B,C,D,E]$ initially. The vertical axis in Fig.~\ref{fig:Dendrogram} represents the distance between clusters.

In hierarchical clustering, we first define a distance function, and all distances between clusters are evaluated. Then, we merge two clusters with the smallest distance. In this paper, we employ the Euclidean distance between a vector $\bm{y}$ and $\bm{z}$:
\begin{align}
    \label{Euclidean}
    L_{\mathrm{E}} &= \sqrt{\sum_{i=1} (y_i - z_i)^2}.
\end{align}

After merging the clusters with the smallest distance, we update the distance between the merged cluster and the other clusters. There are several updating methods, and in this work, we employ the shortest distance method. For example, after merging clusters $A$ and $B$, the updated formula for the distance to cluster $C$ is defined as follows:
\begin{align}
    \label{Single}
    d_{[AB]C} &= \min(d_{AC},d_{BC}),
\end{align}
where $d_{AC}$ and $d_{BC}$ are the distances between clusters $A$ and $C$, and that between clusters $B$ and $C$, respectively. We update the distance between the merged cluster and other clusters for each merging process.

\section{Proposed method}

As mentioned in Sec.~1, since the dictionary size is generally large, the Koopman matrix will also be huge. Therefore, it is desirable to compress the Koopman matrix. Here, we propose a compression method based on hierarchical clustering, which consists of four steps. 

\subsection{Step 1: Compress the Koopman matrix via hierarchical clustering}

\begin{figure}[tb]
\begin{center}
    \includegraphics[width=7cm]{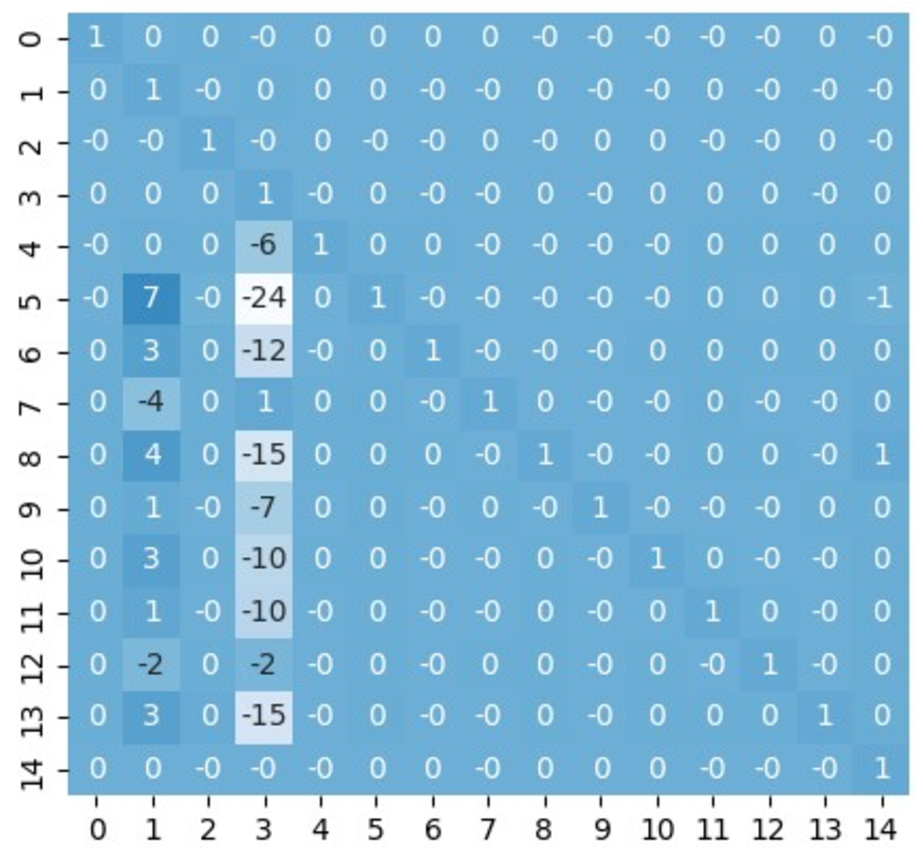}
    \caption{An example of the original matrix for the clustering. Since the Koopman matrix is square, we start from this square matrix in the following explanation.}
    \label{fig:KoopmanMatrix}
\end{center}
\end{figure}

\begin{figure}[tb]
\begin{center}
    \includegraphics[width=7cm]{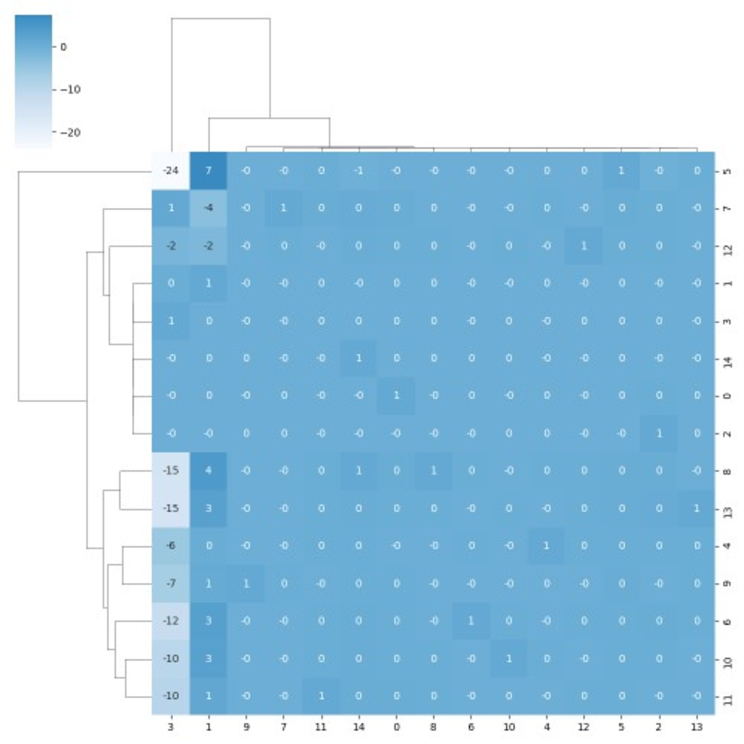}
    \caption{Hierarchical clustering for the matrix. Rows and columns are hierarchically clustered separately.}
    \label{fig:KoopmanClustermap}
\end{center}
\end{figure}

We consider the reduction of the Koopman matrix of Fig.~\ref{fig:KoopmanMatrix} as an example. Hierarchical clustering of this matrix using the shortest distance method with Euclidean distance yields the dendrogram depicted in Fig.~\ref{fig:KoopmanClustermap}. We denote the $N$ clusters in the row direction as $\bm{C}^{\mathrm{r}}=[C^{\mathrm{r}}_1,\cdots,C^{\mathrm{r}}_N]$ and $M$ clusters in the column direction as $\bm{C}^{\mathrm{c}}=[C^{\mathrm{c}}_1,\cdots,C^{\mathrm{c}}_M]$. A cluster consists of some rows or columns; 
$C^{\mathrm{r}}_{ik}$ and $C^{\mathrm{c}}_{jk}$ are the $k$-th components of clusters in $C^{\mathrm{r}}_i$ and $C^{\mathrm{c}}_j$, respectively. The clustering results in Fig.~\ref{fig:KoopmanClustermap} are used to compress the Koopman matrix.

\begin{figure}[tb]
\begin{center}
    \includegraphics[width=9cm]{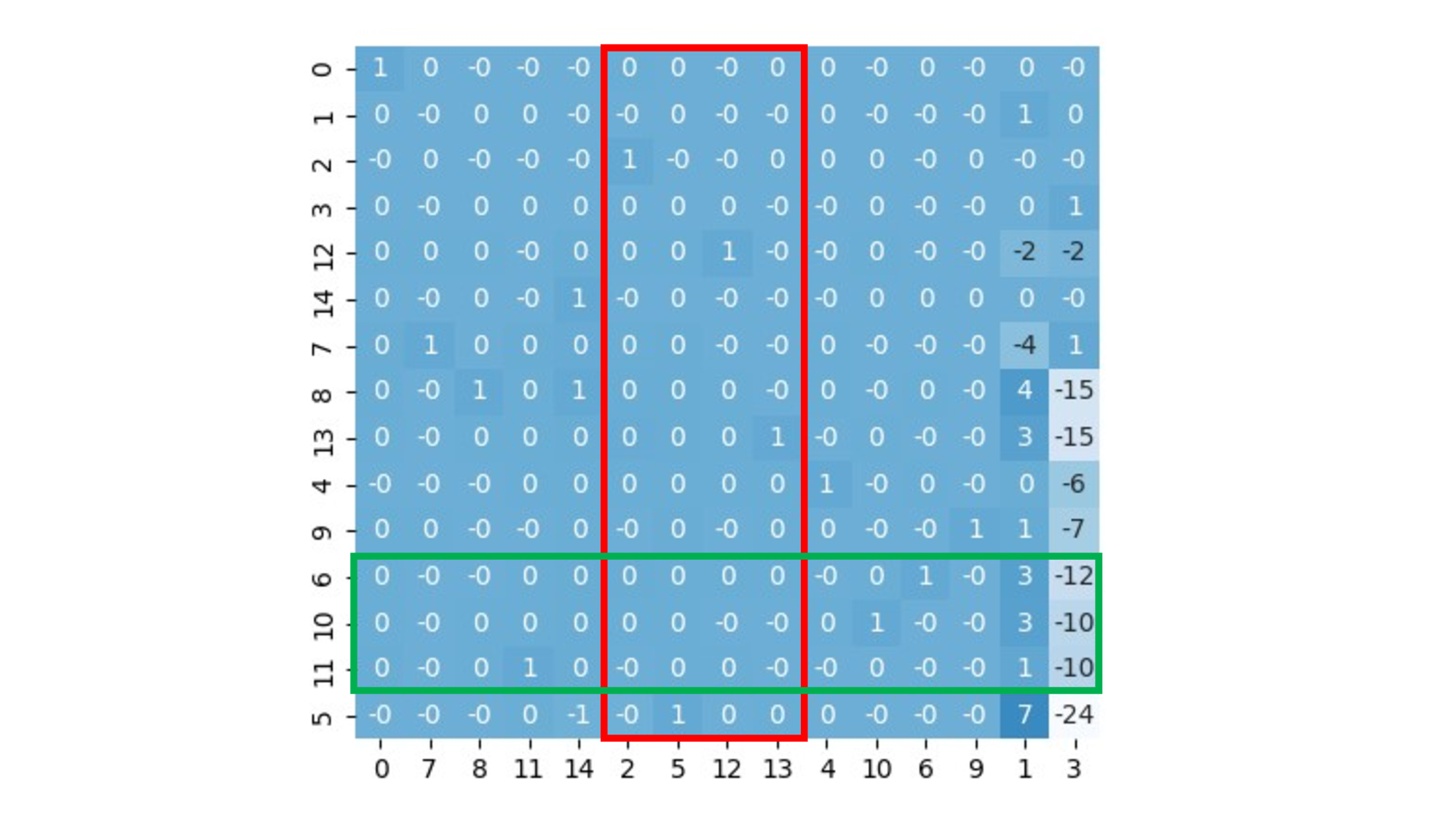}
    \caption{An example of rearranged Koopman matrix by the hierarchical clustering results. The columns surrounded by the red block and the rows surrounded by the green block belong to the same cluster, respectively.}
    \label{fig:KoopmanSort}
\end{center}
\end{figure}

Figure~\ref{fig:KoopmanSort} shows an example of a matrix in which we sort the rows and columns so that elements in the same clusters are adjacent to each other. The green block in the rows and the red one in the columns in Fig.~\ref{fig:KoopmanSort} are examples of rows and columns belonging to the same cluster, respectively. The results of hierarchical clustering yield similar rows or columns. Hence, we assume that the rows or columns in the same cluster have the same values. From this assumption, we set
\begin{align}
    \label{KoopmanRelationRow}
    &K_{C^{\mathrm{r}}_{i1}, j} = K_{C^{\mathrm{r}}_{i2}, j} =\cdots = K_{C^{\mathrm{r}}_{i |C^{\mathrm{r}}_i |}, j},\\
    \label{KoopmanRelationCol}
    &K_{j,C^{\mathrm{c}}_{i1}} = K_{j, C^{\mathrm{c}}_{i2}} =\cdots = K_{j, C^{\mathrm{c}}_{i |C^{\mathrm{c}}_j |}}.
\end{align}
From this assumption, we construct a compressed Koopman matrix by grouping the row-by-row and column-by-column elements in the same cluster. After the grouping procedure, we take the average of elements in the same cluster and set it as the compressed element. That is, the $ij$ element of the compressed Koopman matrix $K^{\prime} \in \mathbb{R}^{N\times M}$ is evaluated as
\begin{align}
    \label{LowKoopman}
    K^{\prime}_{ij} = \frac{\sum^{|C^{\mathrm{r}}_{i}|}_{k=1} \sum^{|C^{\mathrm{c}}_{j}|}_{l=1}  K_{C^{\mathrm{r}}_{ik},C^{\mathrm{c}}_{jl}}}
{|C^{\mathrm{r}}_{i}| |C^{\mathrm{c}}_{j}|}.
\end{align}

\begin{figure}[tb]
\begin{center}
    \includegraphics[width=9cm]{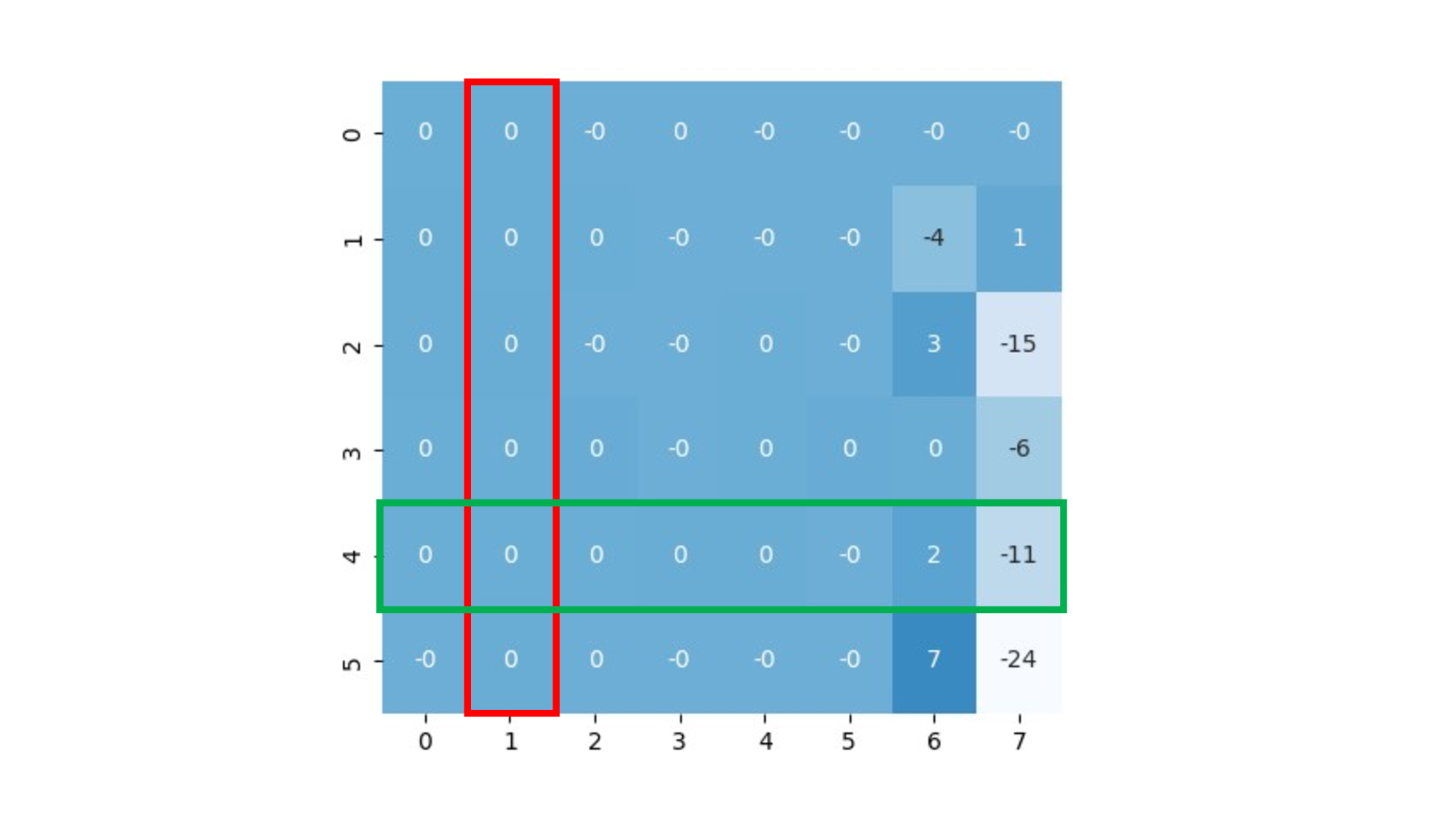}
    \caption{An example of the compressed Koopman matrix based on the proposed method.}
    \label{fig:LowKoopmanmap}
\end{center}
\end{figure}

As an example, let us consider the case with $N=6$ clusters in the row direction and $M=8$ clusters in the column direction. Then, the original Koopman matrix $K \in \mathbb{R}^{15 \times 15}$ is compressed as a matrix $K^\prime \in \mathbb{R}^{6 \times 8}$. Figure~\ref{fig:LowKoopmanmap} shows the compressed Koopman matrix. The green and red blocks in Fig.~\ref{fig:KoopmanSort} correspond to them in Fig.~\ref{fig:LowKoopmanmap}, respectively.

Although the compressed Koopman matrix is derived, we should pay attention to the matrix size. The original Koopman matrix is square. Hence, it is possible to apply the Koopman matrix repeatedly:
\begin{align}
\bm{\psi}(\bm{x}_{t+2}) = K \bm{\psi}(\bm{x}_{t+1}) = K^2 \bm{\psi}(\bm{x}_t)
\end{align}
However, the compressed Koopman matrix $K^\prime$ is not square. Hence, we need different dictionaries for the compressed Koopman matrix. To construct them, we here distinguish dictionaries before and after the matrix acts on them. Let $\bm{\psi}_{\mathrm{B}}(\bm{x}_t)$ and $\bm{\psi}_{\mathrm{A}}(\bm{x}_t)$ be the dictionaries before and after the action, respectively. Then, 
\begin{align}
    K\bm{\psi}_{\mathrm{B}}(\bm{x}_t) = \bm{\psi}_{\mathrm{A}}(\bm{x}_t).
\end{align}
Note that $\bm{\psi}(\bm{x}_{t+1}) = \bm{\psi}_{\mathrm{B}}(\bm{x}_{t+1})=\bm{\psi}_{\mathrm{A}}(\bm{x}_t)$. In the following two steps, we construct two compressed dictionaries, respectively.

\subsection{Step 2: Construct the dictionary after the action}

As an example, we consider a $3\times 3$ Koopman matrix. Then,
\begin{align}
&    \begin{bmatrix}
        K_{11} & K_{12} & K_{13} \\
        K_{21} & K_{22} & K_{23} \\
        K_{31} & K_{32} & K_{33}
    \end{bmatrix}
    \begin{bmatrix}
        \psi_{\mathrm{B}1}(\bm{x}_t) \\ \psi_{\mathrm{B}2}(\bm{x}_t) \\ \psi_{\mathrm{B}3}(\bm{x}_t)
    \end{bmatrix} 
\label{ExampleKoopman1} \nonumber \\
&=
    \begin{bmatrix}
        K_{11}\psi_{\mathrm{B}1}(\bm{x}_t)+K_{12}\psi_{\mathrm{B}2}(\bm{x}_t)+K_{13}\psi_{\mathrm{B}3}(\bm{x}_t) \\
        K_{21}\psi_{\mathrm{B}1}(\bm{x}_t)+K_{22}\psi_{\mathrm{B}2}(\bm{x}_t)+K_{23}\psi_{\mathrm{B}3}(\bm{x}_t) \\
        K_{31}\psi_{\mathrm{B}1}(\bm{x}_t)+K_{32}\psi_{\mathrm{B}2}(\bm{x}_t)+K_{33}\psi_{\mathrm{B}3}(\bm{x}_t)
    \end{bmatrix}\\
\label{ExampleKoopman2}
&=
    \begin{bmatrix}
        \psi_{\mathrm{A}1}(\bm{x}_t) \\ \psi_{\mathrm{A}2}(\bm{x}_t) \\ \psi_{\mathrm{A}3}(\bm{x}_t)
    \end{bmatrix}.
\end{align}

Let us consider the case where the second and third rows are within the same group, i.e. $\bm{C}^{\mathrm{r}}=[[1],[2,3]]$. Then, we construct the matrix elements $K_{2i}^{\prime}$ from $K_{2i}$ and $K_{3i}$. Hence, we have
\begin{align}
    \begin{bmatrix}
        \psi_{\mathrm{A}1}^{\prime}(\bm{x}_t) \\ \psi_{\mathrm{A}2}^{\prime}(\bm{x}_t)
    \end{bmatrix}
=
    \begin{bmatrix}
        K_{11} & K_{12} & K_{13} \\
        K_{21}^{\prime} & K_{22}^{\prime} & K_{23}^{\prime}
    \end{bmatrix}
    \begin{bmatrix}
        \psi_{\mathrm{B}1}(\bm{x}_t) \\ \psi_{\mathrm{B}2}(\bm{x}_t) \\ \psi_{\mathrm{B}3}(\bm{x}_t)
    \end{bmatrix}.
\label{ExampleRowReduce}
\end{align}
We want to construct a compressed dictionary $\psi^{\prime}_{\mathrm{A}}(\bm{x})$ from $\psi_{\mathrm{A}}(\bm{x})$. Note that the second row in Eq.~\eqref{ExampleKoopman1} is a linear combination of $\{K_{2i}\}$ and $\{\psi_{\mathrm{\mathrm{B}}i}\}$. From the relation in Eq.~\eqref{KoopmanRelationRow}, we have $K_{2i} = K_{3i}$. Hence, the second and third rows in Eq.~\eqref{ExampleKoopman1} should also be equal. This fact leads to 
\begin{align}
    \psi_{\mathrm{A}2}^{\prime}(\bm{x}_t) = \psi_{\mathrm{A}2}(\bm{x}_t) = \psi_{\mathrm{A}3}(\bm{x}_t).
\label{eq:comporess_dict_row_equal}
\end{align}
Of course, the grouping procedure based on the clustering results does not give an exact equality. Hence, we approximate $\psi_{\mathrm{A}2}^{\prime}(\bm{x}_t)$ with the average:
\begin{align}
    \psi_{\mathrm{A}2}^{\prime}(\bm{x}_t) = \frac{1}{2} \left( \psi_{\mathrm{A}2}(\bm{x}_t) + \psi_{\mathrm{A}3}(\bm{x}_t) \right).
\end{align}

In summary, we define the compressed dictionary after the action as follows:
\begin{align}
    \psi_{\mathrm{A}i}^{\prime}(\bm{x}_t) = \frac{1}{|C^{\mathrm{r}}_j|}\sum^{|C^{\mathrm{r}}_j|}_{k=1}\psi_{\mathrm{A}k}(\bm{x}_t).
\label{eq:compress_dict_row}
\end{align}

\subsection{Step 3: Construct the dictionary before the action}

Again, we consider the example in Eq.~\eqref{ExampleKoopman1} and Eq.~\eqref{ExampleKoopman2}. When we consider the case with $\bm{C}^{\mathrm{c}}=[[1],[2,3]]$, the following equation should be treated:
\begin{align}
    \label{ExampleColReduce}
    &\begin{bmatrix}
        K_{11} & K_{12}^{\prime}\\
        K_{21} & K_{22}^{\prime}\\
        K_{31} & K_{32}^{\prime}
    \end{bmatrix}
    \begin{bmatrix}
        \psi_{\mathrm{B}1}(\bm{x}_t) \\ \psi_{\mathrm{B}2}^{\prime}(\bm{x}_t)
    \end{bmatrix}
=
    \begin{bmatrix}
        \psi_{\mathrm{A}1}(\bm{x}_t) \\ \psi_{\mathrm{A}2}(\bm{x}_t) \\ \psi_{\mathrm{A}3}(\bm{x}_t)
    \end{bmatrix}.
\end{align}
Here, we want to construct $\psi_{\mathrm{B}2}^{\prime}(\bm{x}_t)$ from $\psi_{\mathrm{B}2}(\bm{x}_t)$ and $\psi_{\mathrm{B}3}(\bm{x}_t)$. The left-hand side of Eq.~\eqref{ExampleColReduce} is calculated as
\begin{align}
    \label{ExampleColReduce2}
\begin{bmatrix}
        K_{11} & K_{12}^{\prime}\\
        K_{21} & K_{22}^{\prime}\\
        K_{31} & K_{32}^{\prime}
\end{bmatrix}
\begin{bmatrix}
        \psi_{\mathrm{B}1}(\bm{x}_t) \\ \psi_{\mathrm{B}2}^{\prime}(\bm{x}_t)
\end{bmatrix}
=
\begin{bmatrix}
        K_{11}\psi_{\mathrm{B}1}(\bm{x}_t)+K_{12}^{\prime}\psi_{\mathrm{B}2}^{\prime}(\bm{x}_t)\\
        K_{21}\psi_{\mathrm{B}1}(\bm{x}_t)+K_{22}^{\prime}\psi_{\mathrm{B}2}^{\prime}(\bm{x}_t)\\
        K_{31}\psi_{\mathrm{B}1}(\bm{x}_t)+K_{32}^{\prime}\psi_{\mathrm{B}2}^{\prime}(\bm{x}_t)
\end{bmatrix},
\end{align}
and the comparison of the right-hand side of Eq.~\eqref{ExampleColReduce2} and Eq.~\eqref{ExampleKoopman1} leads to the following equations:
\begin{align}
    &K_{j1}\psi_{\mathrm{B}1}(\bm{x}_t)+K_{j2}^{\prime}\psi_{\mathrm{B}2}^{\prime}(\bm{x}_t) \nonumber \\
    &=K_{j1}\psi_{\mathrm{B}1}(\bm{x}_t)+K_{j2}\psi_{\mathrm{B}2}(\bm{x}_t)+K_{j3}\psi_{\mathrm{B}3}(\bm{x}_t).
\end{align}
Hence, we have
\begin{align}
    K_{j2}^{\prime}\psi_{\mathrm{B}2}^{\prime}(\bm{x}_t)
    =K_{j2}\psi_{\mathrm{B}2}(\bm{x}_t)+K_{j3}\psi_{\mathrm{B}3}(\bm{x}_t)\label{ExampleRowEqual}.
\end{align}
Using Eq.~\eqref{KoopmanRelationCol} and Eq.~\eqref{LowKoopman}, we have $K_{j2}^{\prime} = K_{j2} = K_{j3}$. Then, we derive the following relationship:
\begin{align}
    \psi_{\mathrm{B}2}^{\prime}(\bm{x}_t)=\psi_{\mathrm{B}2}(\bm{x}_t)+\psi_{\mathrm{B}3}(\bm{x}_t).
\label{eq:step3_mid1}
\end{align}

In summary, we define the compressed dictionary before the action as follows:
\begin{align}
    \psi_{\mathrm{B}j}^{\prime}(\bm{x}_t) = \sum^{|C^{\mathrm{c}}_j|}_{k=1} \psi_{C^{\mathrm{c}}_{jk}}(\bm{x}_t).
\label{eq:compress_dict_col}
\end{align}

\subsection{Step 4: Recover the suitable dictionary for the compressed Koopman matrix}

\begin{figure}[tb]
\begin{center}
    \includegraphics[width=6cm]{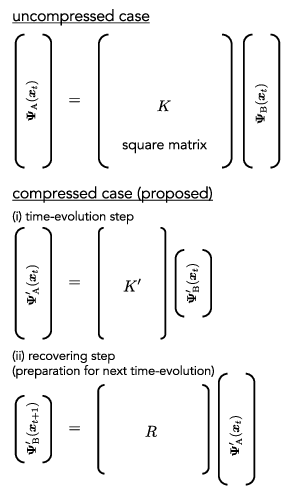}
    \caption{Recovering step. The matrix $R$ recovers the dictionary before the action, $\bm{\psi}_{\mathrm{B}}(\bm{x})$, from the dictionary after the action,  $\bm{\psi}_{\mathrm{A}}(\bm{x})$.}
    \label{fig:recovering}
\end{center}
\end{figure}

The above two steps (Step 2 and Step 3) allow us to apply the compressed Koopman matrix to the compressed dictionaries:
\begin{align}
    K^{\prime}\bm{\psi}^{\prime}_{\mathrm{B}}(\bm{x}_t) = \bm{\psi}^{\prime}_{\mathrm{A}}(\bm{x}_t).
\end{align}
Note that the two dictionaries $\bm{\psi}_{\mathrm{B}}^{\prime}(\bm{x}_t)$ and $\bm{\psi}_{\mathrm{A}}^{\prime}(\bm{x}_t)$ have different sizes. Then, we cannot apply $K^{\prime}$ repeatedly. Hence, we introduce an additional matrix $R$, which recovers the suitable size for the dictionary. Figure~\ref{fig:recovering} shows an image of the procedure. The matrix $R$ acts on $\bm{\Psi}_{\mathrm{A}}^{\prime}(\bm{x})$ as follows:
\begin{align}
    \label{RemakeMatrix}
    R \bm{\psi}_{\mathrm{A}}^{\prime}(\bm{x}_t) = \bm{\psi}_{\mathrm{B}}^{\prime}(\bm{x}_{t+1}).
\end{align}

To explain the matrix $R$, we here use a $5 \times 5$ matrix as an example. Assume that $C^{\mathrm{r}} = [[1,5],[3],[2,4]]$ in the row direction and $C^{\mathrm{c}} = [[1,2],[3,4,5]]$ in the column direction. Hence, we have
\begin{align}
    \begin{bmatrix}
        K_{11}^{\prime} & K_{12}^{\prime} \\
        K_{21}^{\prime} & K_{22}^{\prime} \\
        K_{31}^{\prime} & K_{32}^{\prime} \\
    \end{bmatrix}
    \begin{bmatrix}
        \psi_{\mathrm{B}1}^{\prime}(\bm{x}_t) \\ \psi_{\mathrm{B}2}^{\prime}(\bm{x}_t)
    \end{bmatrix}=
    \begin{bmatrix}
        \psi_{\mathrm{A}1}^{\prime}(\bm{x}_t) \\ \psi_{\mathrm{A}2}^{\prime}(\bm{x}_t) \\ \psi_{\mathrm{A}3}^{\prime}(\bm{x}_t)
    \end{bmatrix}.
\end{align}
Equation~\eqref{eq:compress_dict_col} leads to
\begin{align}
    \psi_{\mathrm{B}1}^{\prime}(\bm{x}_t) &= \psi_1(\bm{x}_t) + \psi_2(\bm{x}_t),\\
    \psi_{\mathrm{B}2}^{\prime}(\bm{x}_t) &= \psi_3(\bm{x}_t) + \psi_4(\bm{x}_t) + \psi_5(\bm{x}_t).
\end{align}
In addition, we here employ the simple equalities as in Eq.~\eqref{eq:comporess_dict_row_equal};
\begin{align}
    \psi_{\mathrm{A}1}^{\prime}(\bm{x}_t) &= \psi_{1}(\bm{x}_{t+1}) = \psi_{5}(\bm{x}_{t+1}),\\
    \psi_{\mathrm{A}2}^{\prime}(\bm{x}_t) &= \psi_{3}(\bm{x}_{t+1}),\\
    \psi_{\mathrm{A}3}^{\prime}(\bm{x}_t) &= \psi_{2}(\bm{x}_{t+1}) = \psi_{4}(\bm{x}_{t+1}).
\end{align}
Since we want to satisfy Eq.~\eqref{RemakeMatrix}, $\bm{\psi}_{\mathrm{B}}(\bm{x}_{t+1})$ must be connected to $\bm{\psi}_{\mathrm{A}}(\bm{x}_{t})$. Then, we derive the following equations from the above relationships:
\begin{align}
    \psi_{\mathrm{B}1}^{\prime}(\bm{x}_{t+1}) &= \psi_{\mathrm{A}1}^{\prime}(\bm{x}_t) + \psi_{\mathrm{A}3}^{\prime}(\bm{x}_t),\\
    \psi_{\mathrm{B}2}^{\prime}(\bm{x}_{t+1}) &= \psi_{\mathrm{A}2}^{\prime}(\bm{x}_t) + \psi_{\mathrm{A}3}^{\prime}(\bm{x}_t) + \psi_{\mathrm{A}1}^{\prime}(\bm{x}_t),
\end{align}
which leads to 
\begin{align}
    \bm{R} = 
    \begin{bmatrix}
        1 & 0 & 1 \\
        1 & 1 & 1
    \end{bmatrix}.
\end{align}

In summary, the matrix for the recovering step, $R$, is constructed as follows:
\begin{align}
    \label{MakeRemakeMatrix}
    R_{ji} = \sum^{|C^{\mathrm{r}}_{i}|}_{k=1} \sum^{|C^{\mathrm{c}}_{j}|}_{l=1} \1 [C^{\mathrm{r}}_{ik}=C^{\mathrm{c}}_{jl}]
\end{align}
where $\1$ is the indicator function, which returns $1$ when the condition is satisfied and $0$ if not.

\subsection{Remark on the procedure}
\label{LowKoopmanandRemakeMatrix}

There are two ways for the usage of the matrix $R$;
\begin{align}
    R K^{\prime} \bm{\psi}_{\mathrm{B}}(\bm{x}_t) &= K_{\mathrm{B}}^{\prime} \bm{\psi}_{\mathrm{B}}(\bm{x}_t) = \bm{\psi}_{\mathrm{B}}(\bm{x}_{t+1}),\label{VB}\\
    K^{\prime} R \bm{\psi}_A(\bm{x}_t) &= K_{\mathrm{A}}^{\prime} \bm{\psi}_{\mathrm{A}}(\bm{x}_t) = \bm{\psi}_{\mathrm{A}}(\bm{x}_{t+1}).\label{VA}
\end{align}
Equation~\eqref{VB} is a time evolution using the dictionary before the action, $\bm{\psi}_{\mathrm{B}}(\bm{x}_{t})$; Eq.~\eqref{VA} is that for $ \bm{\psi}_{\mathrm{A}}(\bm{x}_{t})$. Note that these two procedures employ matrices with different size, i.e., $R K^{\prime} = K^{\prime}_{\mathrm{B}} \in \mathbb{R}^{M\times M}$ and $K^{\prime} R = K^{\prime}_{\mathrm{A}} \in \mathbb{R}^{N\times N}$.

\section{Numerical experiments}

In this section, we demonstrate the proposed method using the concrete physical model. We here use a little difficult model for the prediction tasks, i.e., the cart-pole model controlled by a reinforcement learning model. The cart-pole model aims to stand the pole only with force on the cart. The underlying system equations of the cart-pole model are nonlinear. Of course, there is no need to know the underlying system equations; we only use a data set for the dynamics. Here, we construct the Koopman matrix by the EDMD algorithm and evaluate the features of the compressed Koopman matrix.

\subsection{Data generation}

We artificially generate the data for the cart-pole model. The cart-pole model is denoted in Sec.~2.1 and Fig.~\ref{fig:cart_pole}. 

The details of the data generation is as follows. The initial state is set as $\bm{x}_0 = [0,0,0,0]$. The time-evolution is simulated with $\Delta t = 0.05$, and the final time is set as $T = 5$. Hence, we obtain a data set with $100$ snapshot pairs from a single trajectory data. In the numerical experiments, we apply the control inputs obtained from the pre-trained Q-learning model so that the pole stands. This procedure is repeated $100$ times to generate the training data set; we finally obtain a training data set with $10,000$ snapshot pairs. Furthermore, we employ further $100$ repeated simulations with the same settings to obtain the evaluation data set.

\subsection{Definition of compression ratio}

We define the compression ratio as the ratio of the row and column size of $K^{\prime}$ to those of $K$. The size of $K^{\prime}$ is $N \times M$, and that of $K$ is $D \times D$. Hence, we denote the ratio as $[N/D, M/D]$. Note that the ratio $[1.0,1.0]$ corresponds to the uncompressed case. We use the monomial dictionary functions with  a total degree of up to $10$, and then $D=1,001$.

In Sec.~\ref{LowKoopmanandRemakeMatrix}, we mentioned two types of formulas, i.e., Eq.~\eqref{VA} and Eq.~\eqref{VB}. In the following numerical experiments, we employ Eq.~\eqref{VA}. The dictionary before the action used in Eq.~\eqref{VB} is constructed from the sum of the original dictionary functions; see Eq.~\eqref{eq:step3_mid1}. Hence, we require accurate evaluations for all these functions. By contrast, the dictionary after the action used in Eq.~\eqref{VA} is based on the relation of equality; see Eq.~\eqref{eq:comporess_dict_row_equal}. Hence, we expect the numerical evaluation of Eq.~\eqref{VA} to be more stable than that of Eq.~\eqref{VB}. From some preliminary numerical experiments, we determined the use of Eq.~\eqref{VA}.

\subsection{Computation time}

First, we compare the difference in computation time by compression of the Koopman matrix. Here, we output the average computation time per step for the evaluation data, i.e., 10,000 steps. The average computation time per step is shown in Table~\ref{ComputeAverage}. The row shows the row size ratio of the compressed Koopman matrix, and the column corresponds to the column size ratio.

\begin{table}[bt]
    \caption{Average computation time per step [ms]. The rows and columns correspond to different row size ratios and column size ratios, respectively.}
    \label{ComputeAverage}
    \centering
    \begin{tabular}{c|ccccc}
        \hline
        Row $\backslash$ Column & $1.0$ & $0.8$ & $0.6$ & $0.4$ & $0.2$\\
        \hline
        $1.0$  & 0.524 & 0.529 & 0.558 & 0.490 & 0.243 \\
        $0.8$  & 0.404 & 0.392 & 0.349 & 0.345 & 0.143 \\
        $0.6$  & 0.192 & 0.176 & 0.151 & 0.141 & 0.123 \\
        $0.4$  & 0.120 & 0.098 & 0.080 & 0.098 & 0.116 \\
        $0.2$  & 0.072 & 0.057 & 0.106 & 0.060 & 0.063 \\
        \hline
    \end{tabular}
\end{table}

Table~\ref{ComputeAverage} indicates that we can reduce the computation time largely by compressing in the row direction. The reason is that the row size ratio determines the size of the square matrix $K^{\prime}_{\mathrm{A}}$. Hence, the small row size ratio yields a small square matrix $K^{\prime}_{\mathrm{A}}$, and we achieve a large decrease in the computation time. 

\subsection{Prediction accuracy}

\begin{figure}[tb]
    \centering
    \includegraphics[width=8cm]{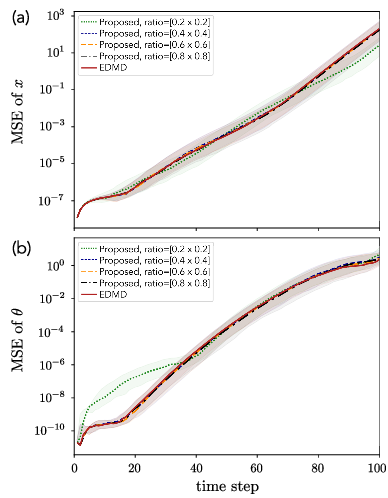}
    \caption{Mean squared errors between the predictions and the true values in the evaluation data set. (a) Results for the cart coordinate $x$. (b) Results for the pole angle $\theta$. The colored areas represent the corresponding quartiles.}
    \label{fig:ResultX}
\end{figure}

While the computation time is reduced, we must confirm that the compressed Koopman matrix retains accuracy even with the reduction in dimensions. Here, we compare the differences between the predictions and the true values in the evaluation data set. Figure~\ref{fig:ResultX} shows the mean squared errors of the cart coordinate $x$ and the pole angle $\theta$ for the original Koopman matrix and the compressed ones of size ratios $[0.8, 0.8]$, $[0.6, 0.6]$, $[0.4, 0.4]$, and $[0.2, 0.2]$. Note that the accuracy for the cart position $x$ is not good even in the uncompressed original Koopman matrix. This is because the cart-pole model aims to stand the poles, which causes the cart to move significantly. Actually, the cart position largely varied in the trajectory data controlled by the reinforcement learning model. Hence, it is difficult to predict the position of the cart. By contrast, the pole angle variations are smaller due to the control, which leads to better predictive accuracy. From Fig.~\ref{fig:ResultX}, one may consider that a long-term prediction is difficult. However, it is general to perform the control based on short-term prediction while observing the system state. Hence, this length of time will be sufficient. Of course, the aim here is not to make the control inputs; we investigate the effects of the compression of the Koopman matrix on the prediction accuracy. Then, it is enough to compare the accuracy of the proposed methods with the uncompressed original Koopman matrix.

The results in Fig.~\ref{fig:ResultX} indicate that the accuracies for the cases with $[0.8, 0.8]$, $[0.6, 0.6]$, $[0.4, 0.4]$ are comparable to that of the conventional method. For the case with $[0.2, 0.2]$, the results show different behaviors; for $x$, the mean squared error is smaller than that of the conventional method in the large time steps, while the result is worse for $\theta$. Hence, the ratio $[0.2, 0.2]$ is not enough to recover the uncompressed results. From these results, we confirm that the proposed method works well even in large compression with $[0.4, 0.4]$. As we saw in Sec.~4.3, the $[0.4,0.4]$ case yields a speedup of about $5$ times.

\subsection{Comparison with SVD}

Here, we compare the proposed method with the SVD method, which leads to a low-rank approximation and the corresponding small memory size. 

The usage of the SVD method for Koopman operators has already been discussed in \cite{Dawson2016}. However, the discussion focused on the reduction of computational complexity in the derivation of the Koopman matrix by the EDMD algorithm and the effect of noise in the data. Here, we focus on the SVD and the proposed method in terms of memory reduction of the Koopman matrix. Therefore, we compare the accuracy of the results of the SVD method with the results of the compressed Koopman matrix with the comparable memory size.

The SVD yields the following low-rank approximation with $r \leq D$ of the Koopman matrix:
\begin{align}
  K &\simeq K_U^r \bm{K}_{\Sigma}^r (K_V^r)^{\top} = K_{U\Sigma}^r (K_V^r)^{\top},
\end{align}
where $K_U^r \in \mathbb{R}^{D\times r}$, $K_{\Sigma}^r \in \mathbb{R}^{r\times r}$, $K_V^r \in \mathbb{R}^{D\times r}$, and $K_U^r K_{\Sigma}^r = K_{U\Sigma}^r \in \mathbb{R}^{D\times r}$. By using $K_V^r$ and $K_{U\Sigma}^r$, it is possible to construct the corresponding Koopman matrix with the memory-size reduction. Note that the condition $r<\frac{D}{2}$ is necessary to compress the matrix size more than the original matrix size.

The number of elements in the original Koopman matrix is $1,002,001$. Table~\ref{table:ClusteringSize} shows the numbers of matrix elements of $K^{\prime}_{\mathrm{A}}$ for various settings. In Table~\ref{table:SVDSize}, we show the total number of matrix elements for $K_{U\Sigma}^r$ and $(K_V^r)^{\top}$. From these results, we compare the results for $[0.2, 0.2]$ with the $\mathrm{rank}=20$ case and those for $[0.3, 0.3]$ with the $\mathrm{rank}=50$ case.

\begin{table}[bt]
    \begin{center}
    \caption{The number of matrix elements in the proposed method.}
    \label{table:ClusteringSize}
    \begin{tabular}{lc}
        \hline
        Row size ratio & Number of matrix elements\\
        \hline
        0.8  & 640000\\
        0.6  & 360000\\
        0.4  & 160000\\
        0.3  & 90000\\
        0.2  & 40000\\
        \hline
    \end{tabular}
    \end{center}
\end{table}

\begin{table}[bt]
    \begin{center}
    \caption{The total number of matrix elements for $K_{U\Sigma}^r$ and $(K_V^r)^{\top}$ in the SVD method.}
    \label{table:SVDSize}
    \begin{tabular}{lc}
        \hline
        Rank & Number of matrix elements\\
        \hline
        300  & 600600\\
        200  & 400400\\
        100  & 200200\\
        50  & 100100\\
        20  & 40040\\
        \hline
    \end{tabular}
    \end{center}
\end{table}

\begin{figure}[t]
\begin{center}
    \includegraphics[width=8cm]{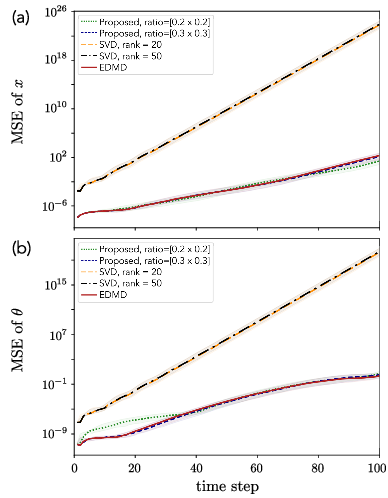}
\end{center}
    \caption{Mean squared errors between the predictions and the true values in the evaluation data set. The results obtained by the SVD method are included. (a) Results for the cart coordinate $x$. (b) Results for the pole angle $\theta$. The colored areas represent the corresponding quartiles.}
    \label{fig:ResultXSVD}
\end{figure}

Figure~\ref{fig:ResultXSVD} shows the results. We see that the low-rank approximation via the SVD gives worse predictions than the proposed method. Hence, we conclude that the proposed method reduces the memory size while maintaining accuracy better than the low-rank approximation of the SVD.

These result suggests that the hierarchical structure could extract and preserve some important structures for physical simulation models.

\section{Conclusion}

The large dictionary size yields the large Koopman matrix, which results in considerably high computational costs. To reduce the computational cost, we confirm that the compression method based on hierarchical clustering is more effective than the conventional SVD method. The proposed method based on hierarchical clustering could effectively employ the information embedded in the Koopman matrix.

Of course, the adequate size ratio of the compressed Koopman matrix could be different for other physical models and settings. In practice, we should consider methods to find the optimal size of the compressed Koopman matrices. In addition, we will pursue an interpretation of the results by hierarchical clustering from theoretical interests. At this stage, we cannot find a useful interpretation, and more extensive studies will be necessary in the future. 

While the usage of the Koopman operator theory has been actively studied recently, there is room to discuss the features of the Koopman matrix. The present work is the first attempt to apply hierarchical clustering to this research topic. We believe that this idea will be helpful for future research.

\begin{acknowledgments}
This work was supported by JST FOREST Program (Grant Number JPMJFR216K, Japan).
\end{acknowledgments}

\end{document}